\documentclass{article}

\usepackage{fullpage,times,hyperref}
\usepackage{graphicx}
\usepackage{subfigure}
\usepackage{algorithm,algorithmic}

\usepackage{color}
\usepackage{amsmath,amsthm,amssymb}

\newtheorem{theorem}{Theorem}
\newtheorem{lemma}[theorem]{Lemma}

\newtheorem{remark}[theorem]{Remark}

\newenvironment{definition}[1][Definition.]{\begin{trivlist}
\item[\hskip \labelsep {\bfseries #1}]}{\end{trivlist}}

\def\wh{\widehat}
\def\wt{\widetilde}

\def\X{\mathcal{X}}
\def\Y{\mathcal{Y}}
\def\A{\mathcal{A}}
\def\R{\mathbb{R}}
\def\E{\mathbb{E}}
\def\N{\mathbb{N}}
\def\rec{\mathsf{R}}
\def\learn{\mathsf{L}}

\DeclareMathOperator{\sperr}{sperr}
\DeclareMathOperator{\supp}{supp}

\DeclareMathOperator{\trace}{trace}

\title{Multi-Label Prediction via Compressed Sensing}

\author{
Daniel Hsu\thanks{UC San Diego, e-mail: {\tt djhsu@cs.ucsd.edu}}
\and Sham M.~Kakade\thanks{TTI-Chicago, e-mail: {\tt sham@tti-c.org}}
\and John Langford\thanks{Yahoo!~Research, e-mail: {\tt jl@hunch.net}}
\and Tong Zhang\thanks{Rutgers University, e-mail: {\tt tongz@rci.rutgers.edu}}
}

\begin{document}

\maketitle

\begin{abstract}
We consider multi-label prediction problems with large output spaces under
the assumption of \emph{output sparsity} -- that the target (label) vectors
have small support.
We develop a general theory for a variant of the popular error correcting
output code scheme, using ideas from compressed sensing for exploiting this
sparsity.
The method can be regarded as a simple reduction from multi-label
regression problems to binary regression problems.
We show that the number of subproblems need only be logarithmic in the
total number of possible labels, making this approach radically more
efficient than others.
We also state and prove robustness guarantees for this method in the form
of regret transform bounds (in general), and also provide a more detailed
analysis for the linear prediction setting.
\end{abstract}

\section{Introduction}
\label{sec:introduction}

Suppose we have a large database of images, and we want to learn to predict
who or what is in any given one.
A standard approach to this task is to collect a sample of these images
$x$ along with corresponding labels $y = (y_1,\ldots,y_d) \in \{0,1\}^d$,
where $y_i = 1$ if and only if person or object $i$ is depicted in image
$x$, and then feed the labeled sample to a multi-label learning algorithm.
Here, $d$ is the total number of entities depicted in the entire database.
When $d$ is very large (\emph{e.g.}~$10^3$, $10^4$), the simple
one-against-all approach of learning a single predictor for each entity can
become prohibitively expensive, both at training and testing time.

Our motivation for the present work comes from the observation that
although the output (label) space may be very high dimensional, the
actual labels are often sparse.
That is, in each image, only a small number of entities may be present and
there may only be a small amount of ambiguity in who or what they are.
In this work, we consider how this sparsity in the output space, or
\emph{output sparsity}, eases the burden of large-scale multi-label
learning.

\noindent {\bf Exploiting output sparsity.}
A subtle but critical point that distinguishes output sparsity from more
common notions of sparsity (say, in feature or weight vectors)
is that we are interested in the sparsity of $\E[y|x]$ rather than $y$.
In general, $\E[y|x]$ may be sparse while the actual outcome $y$ may not
(\emph{e.g.}~if there is much unbiased noise); and, vice versa, $y$ may be
sparse with probability one but $\E[y|x]$ may have large support
(\emph{e.g.}~if there is little distinction between several labels).

Conventional linear algebra suggests that we must predict $d$ parameters in
order to find the value of the $d$-dimensional vector $\E[y|x]$ for each
$x$.
A crucial observation -- central to the area of compressed sensing
\cite{donoho06} -- is that methods exist to recover $\E[y|x]$ from just
$O(k \log d)$ measurements when $\E[y|x]$ is $k$-sparse.
This is the basis of our approach.

\noindent {\bf Our contributions.} We show how to apply algorithms for
compressed sensing to the output coding approach \cite{db95}.  At a
high level, the output coding approach creates a collection of
subproblems of the form ``Is the label in this subset or its
complement?'', solves these problems, and then uses their solution to
predict the final label.

The role of compressed sensing in our application is distinct from its more
conventional uses in data compression.
Although we do employ a sensing matrix to compress training data, we
ultimately are not interested in recovering data explicitly compressed
this way.
Rather, we \emph{learn to predict compressed label vectors}, and then use
sparse reconstruction algorithms to \emph{recover uncompressed labels from
these predictions}.
Thus we are interested in reconstruction accuracy of predictions, averaged
over the data distribution.

The main contributions of this work are:
\begin{enumerate}
\item A formal application of compressed sensing to prediction
  problems with output sparsity.

\item An efficient output coding method, in which the number of required
  predictions is only logarithmic in the number of labels $d$, making it
  applicable to very large-scale problems.


\item Robustness guarantees, in the form of regret transform bounds (in
general) and a further detailed analysis for the linear prediction setting.

\end{enumerate}

%

\noindent {\bf Prior work.}
The ubiquity of multi-label prediction problems in domains ranging from
multiple object recognition in computer vision to automatic keyword tagging
for content databases has spurred the development of numerous general
methods for the task.
Perhaps the most straightforward approach is the well-known one-against-all
reduction~\cite{rk04}, but this can be too expensive when the number of
possible labels is large (especially if applied to the power set of the
label space \cite{blsb04}).
When structure can be imposed on the label space (\emph{e.g.}~class
hierarchy), efficient learning and prediction methods are often
possible~\cite{ck01,tck03,cbgtz04,thja04,rssst06}.
Here, we focus on a different type of structure, namely output sparsity,
which is not addressed in previous work.
Moreover, our method is general enough to take advantage of structured
notions of sparsity (\emph{e.g.}~group sparsity) when
available~\cite{hzm09}.
Recently, heuristics have been proposed for discovering structure in large
output spaces that empirically offer some degree of
efficiency~\cite{tkv08}.

As previously mentioned, our work is most closely related to the class of
output coding method for multi-class prediction, which was first introduced
and shown to be useful experimentally in \cite{db95}.
Relative to this work, we expand the scope of the approach to multi-label
prediction and provide bounds on regret and error which guide the design of
codes.  
The loss based decoding approach~\cite{Singer} suggests decoding so as
to minimize loss.
However, it does not provide significant guidance in the choice of encoding
method, or the feedback between encoding and decoding which we analyze
here.

The output coding approach is inconsistent when classifiers are used
and the underlying problems being encoded are noisy.
This is proved and analyzed in \cite{lb05}, where it is also shown that
using a Hadamard code creates a robust consistent predictor when reduced to
binary regression.
Compared to this method, our approach achieves the same robustness
guarantees up to a constant factor, but requires training and evaluating
exponentially (in $d$) fewer predictors.

Our algorithms rely on several methods from compressed sensing, which we
detail where used.

\section{Preliminaries}
\label{sec:preliminaries}

Let $\X$ be an arbitrary input space and $\Y \subset \R^d$ be a
$d$-dimensional output (label) space.
We assume the data source is defined by a fixed but unknown distribution
over $\X \times \Y$.
Our goal is to learn a predictor $F:\X\to\Y$ with low expected
$\ell_2^2$-error
$\E_x \|F(x) - \E[y|x]\|_2^2$
(the sum of mean-squared-errors over all labels)
using a set of $n$ training data $\{ (x_i,y_i) \}_{i=1}^n$.

We focus on the regime in which the output space is very high-dimensional
($d$ very large), but for any given $x \in \X$, the expected value
$\E[y|x]$ of the corresponding label $y \in \Y$ has only a few non-zero
entries.
A vector is \emph{$k$-sparse} if it has at most $k$ non-zero entries.

%

\section{Learning and Prediction}
\label{sec:learning-and-prediction}

\subsection{Learning to Predict Compressed Labels}
\label{subsec:learning}

Let $A:\R^d \to \R^m$ be a linear compression function, where $m \leq d$
(but hopefully $m \ll d$).
We use $A$ to compress (\emph{i.e.}~reduce the dimension of) the labels
$\Y$, and learn a predictor $H:\X \to A(\Y)$ of these compressed labels.
Since $A$ is linear, we simply represent $A \in \R^{m \times d}$ as a
matrix.

Specifically, given a sample $\{(x_i,y_i)\}_{i=1}^n$, we form a compressed
sample $\{(x_i,Ay_i)\}_{i=1}^n$ and then learn a predictor $H$ of
$\E[Ay|x]$ with the objective of minimizing the $\ell_2^2$-error
$\E_x \|H(x) - \E[Ay|x]\|_2^2$.
%
%


\subsection{Predicting Sparse Labels}
\label{subsec:prediction}

To obtain a predictor $F$ of $\E[y|x]$, we compose the predictor $H$ of
$\E[Ay|x]$ (learned using the compressed sample) with a reconstruction
algorithm $\rec:\R^m\to\R^d$.
The algorithm $\rec$ maps predictions of compressed labels $h \in \R^m$ to
predictions of labels $y \in \Y$ in the original output space.
These algorithms typically aim to find a sparse vector $y$ such that $Ay$
closely approximates $h$.

Recent developments in the area of compressed sensing have produced a spate
of reconstruction algorithms with strong performance guarantees when the
compression function $A$ satisfies certain properties.
We abstract out the relevant aspects of these guarantees in the following
definition.
\begin{definition}
An algorithm $\rec$ is a \emph{valid reconstruction algorithm for a family
of compression functions $( \A_k \subset \bigcup_{m \geq 1} \R^{m \times d} : k \in \N )$
and sparsity error $\sperr:\N \times \R^d \to \R$}, if there exists a
function $f:\N\to\N$ and constants $C_1, C_2 \in \R$ such that: on input $k
\in \N$, $A \in \A_k$ with $m$ rows, and $h \in \R^{m}$, the algorithm
$\rec(k,A,h)$ returns an $f(k)$-sparse vector $\wh{y}$ satisfying
\begin{equation*}
\|\wh{y} - y\|_2^2
\ \leq \
C_1 \cdot \|h - Ay\|_2^2 + C_2 \cdot \sperr(k,y)
\end{equation*}
for all $y \in \R^d$.
The function $f$ is the \emph{output sparsity} of $\rec$ and the constants
$C_1$ and $C_2$ are the \emph{regret factors}.
\end{definition}
Informally, if the predicted compressed label $H(x)$ is close to $\E[Ay|x]
= A\E[y|x]$, then the sparse vector $\wh y$ returned by the reconstruction
algorithm should be close to $\E[y|x]$; this latter distance $\|\wh y -
\E[y|x]\|_2^2$ should degrade gracefully in terms of the accuracy of $H(x)$
and the sparsity of $\E[y|x]$.
Moreover, the algorithm should be agnostic about the sparsity of $\E[y|x]$
(and thus the sparsity error $\sperr(k, \E[y|x])$), as well as the
``measurement noise'' (the prediction error $\|H(x) - \E[Ay|x]\|_2$).
This is a subtle condition and precludes certain reconstruction
algorithm (\emph{e.g.}~Basis Pursuit~\cite{crt06}) that require the user to
supply a bound on the measurement noise.
However, the condition is needed in our application, as such bounds on the
prediction error (for each $x$) are not generally known beforehand.


We make a few additional remarks on the definition.
\begin{enumerate}
\item The minimum number of rows of matrices $A \in \A_k$ may in general
depend on $k$ (as well as the ambient dimension $d$).
In the next section, we show how to construct such $A$ with close to the
optimal number of rows.
\item The sparsity error $\sperr(k,y)$ should measure how poorly $y \in
\R^d$ is approximated by a $k$-sparse vector.
\item A reasonable output sparsity $f(k)$ for sparsity level $k$ should not
be much more than $k$, \emph{e.g.}~$f(k) = O(k)$.
\end{enumerate}

Concrete examples of valid reconstruction algorithms (along with the
associated $\A_k$, $\sperr$, etc.) are given in the next section.

%

\section{Algorithms}
\label{sec:algorithms}

Our prescribed recipe is summarized in Algorithms~\ref{alg:training} and
\ref{alg:prediction}.
We give some examples of compression functions and reconstruction
algorithms in the following subsections.

\begin{figure}
\begin{tabular}[p]{cc}
\begin{minipage}[t]{0.45\textwidth}
\begin{algorithm}[H]
\caption{Training algorithm}
\label{alg:training}
\begin{algorithmic}
  \PARAMETERS sparsity level $k$,
  compression function $A \in \A_k$ with $m$ rows,
  regression learning algorithm $\learn$
  \INPUT training data $S \subset \X \times \R^d$
  \FOR{$i = 1, \ldots, m$}
    \STATE $h_i \gets \learn( \{ (x, (Ay)_i) : (x, y) \in S \} )$
  \ENDFOR
  \OUTPUT regressors $H = [ h_1, \ldots, h_m ]$
\end{algorithmic}
\end{algorithm}
\end{minipage}
&
\begin{minipage}[t]{0.45\textwidth}
\begin{algorithm}[H]
\caption{Prediction algorithm}
\label{alg:prediction}
\begin{algorithmic}
  \PARAMETERS sparsity level $k$,
  compression function $A \in \A_k$ with $m$ rows,
  valid reconstruction algorithm $\rec$ for $\A_k$
  \INPUT regressors $H = [ h_1, \ldots, h_m ]$,
  test point $x \in \X$
  \OUTPUT $\wh{y} = \vec{\rec}(k,A,[ h_1(x), \ldots, h_m(x) ])$
\end{algorithmic}
\end{algorithm}
\end{minipage}
\end{tabular}
\caption{Training and prediction algorithms.}
\end{figure}

\subsection{Compression Functions}
\label{subsec:compression-functions}

Several valid reconstruction algorithms are known for compression matrices
that satisfy a \emph{restricted isometry property}.
\begin{definition}
A matrix $A \in \R^{m \times d}$ satisfies the
\emph{$(k,\delta)$-restricted isometry property ($(k,\delta)$-RIP)},
$\delta \in (0,1)$, if
$(1 - \delta) \|x\|_2^2 \leq \|Ax\|_2^2 \leq (1 + \delta) \|x\|_2^2$
for all $k$-sparse $x \in \R^d$.
\end{definition}
While some explicit constructions of $(k,\delta)$-RIP matrices are known
(\emph{e.g.}~\cite{devore07}), the best guarantees are obtained when the
matrix is chosen randomly from an appropriate distribution, such as one of
the following \cite{mptj08,rv06}.
\begin{itemize}
\item All entries i.i.d.~Gaussian $N(0,1/m)$, with $m = O(k \log (d/k))$.
\item All entries i.i.d.~Bernoulli $B(1/2)$ over $\{\pm1/\sqrt{m}\}$, with
$m = O(k \log (d/k))$.
\item $m$ randomly chosen rows of the $d \times d$ Hadamard matrix over
$\{\pm1/\sqrt{m}\}$, with $m = O(k \log^5 d)$.
\end{itemize}
The hidden constants in the big-$O$ notation depend inversely on $\delta$
and the probability of success.

A striking feature of these constructions is the very mild dependence
of $m$ on the ambient dimension $d$. This translates to a significant
savings in the number of learning problems one has to solve after
employing our reduction.


Some reconstruction algorithms require a stronger guarantee of bounded
\emph{coherence} $\mu(A) \leq O(1/k)$, where $\mu(A)$ defined as
$$ \mu(A) = \max_{1 \leq i < j \leq d} |(A^\top A)_{i,j}|/\sqrt{|(A^\top
A)_{i,i}| |(A^\top A)_{j,j}| } $$
It is easy to check that the Gaussian, Bernoulli, and Hadamard-based random
matrices given above have coherence bounded by $O(\sqrt{(\log d)/m})$ with
high probability.
Thus, one can take $m = O(k^2\log d)$ to guarantee $1/k$ coherence.
This is a factor $k$ worse than what was needed for $(k,\delta)$-RIP, but
the dependence on $d$ is still small.

\subsection{Reconstruction Algorithms}
\label{subsec:reconstruction-algorithms}

In this section, we give some examples of valid reconstruction algorithms.
Each of these algorithm is valid with respect to the sparsity error given
by
\begin{equation*}
\sperr(k,y) \ = \ \|y - y_{(1:k)}\|_2^2 + \frac{1}{k} \|y - y_{(1:k)}\|_1^2
\end{equation*}
where $y_{(1:k)}$ is the best $k$-sparse approximation of $y$
(\emph{i.e.}~the vector with just the $k$ largest (in magnitude)
coefficients of $y$).

The following theorem relates reconstruction quality to approximate sparse
regression, giving a sufficient condition for any algorithm to be valid for
RIP matrices.
\begin{theorem} \label{thm:recon}
Let $\A_k= \{\text{$(k+f(k),\delta)$-RIP matrices}\}$ for some
function $f:\N\to\N$, and let $A \in \A_k$ have $m$ rows.
If for any $h \in \R^m$, a reconstruction algorithm $\rec$ returns an
$f(k)$-sparse solution $\wh{y}=\rec(k,A,h)$ satisfying
  \[
  \|A \wh{y} - h\|_2^2 
  \leq \inf_{y \in \R^d} C \| A y_{(1:k)} - h \|_2^2 ,
  \]
  then it is a valid reconstruction algorithm for $\A_k$ and $\sperr$ given
  above, with output sparsity $f$ and regret factors
  $C_1=2(1+\sqrt{C})^2/(1-\delta)$ and 
  $C_2=4(1+(1+\sqrt{C})/(1-\delta))^2$.
\end{theorem}
Proofs are deferred to Section~\ref{sec:proofs}.

{\bf Iterative and greedy algorithms.}
Orthogonal Matching Pursuit (OMP) \cite{mz93}, FoBa \cite{zhang08}, and CoSaMP
\cite{nt08} are examples of iterative or greedy reconstruction algorithms.
OMP is a greedy forward selection method that repeatedly selects a new
column of $A$ to use in fitting $h$
(see Algorithm~\ref{alg:omp}).
FoBa is similar, except it also incorporates backward steps to un-select
columns that are later discovered to be unnecessary.
CoSaMP is also similar to OMP, but instead selects larger sets of columns
in each iteration.

\begin{figure}
\begin{minipage}[t]{\textwidth}
\begin{algorithm}[H]
  \caption{Prediction algorithm with $\rec = \text{OMP}$}
  \label{alg:omp}
\begin{algorithmic}
  \PARAMETERS sparsity level $k$,
  compression function $A = [a_1| \ldots| a_d] \in \A_k$ with $m$ rows,
  \INPUT regressors $H = [ h_1, \ldots, h_m ]$,
  test point $x \in \X$
  \STATE $h \gets [ h_1(x), \ldots, h_m(x) ]^\top$
  \quad (predict compressed label vector)
  \STATE $\wh{y} \gets \vec{0}$, $J \gets \emptyset$, $r \gets h$
  \FOR{$i = 1, \ldots, 2k$}
    \STATE $j_* \gets \arg\max_j |r^\top a_j| / \|a_j\|_2$
    \quad (column of $A$ most correlated with residual $r$)
    \STATE $J \gets J \cup \{ j_* \}$
    \quad (add $j_*$ to set of selected columns)
    \STATE $\wh{y}_J \gets (A_J)^\dag h$, $\wh{y}_{J^c} \gets \vec{0}$
    \quad (least-squares restricted to columns in $J$)
    \STATE $r \gets h - A\wh{y}$
    \quad (update residual)
  \ENDFOR
  \OUTPUT $\wh{y}$
\end{algorithmic}
\end{algorithm}
\end{minipage}
\caption{Prediction algorithm specialized with Orthogonal Matching Pursuit.}
\end{figure}

FoBa and CoSaMP are valid reconstruction algorithms for RIP matrices
($(8k,0.1)$-RIP and $(4k,0.1)$-RIP, respectively) and have linear output
sparsity ($8k$ and $2k$).
These guarantees are apparent from the cited references.
For OMP, we give the following guarantee.
\begin{theorem} \label{thm:omp}
  If $\mu(A) \leq 0.1/k$, then
  after $f(k)=2k$ steps of OMP, the algorithm returns $\wh{y}$ satisfying
  \[
  \|A \wh{y} - h\|_2^2 \leq 23
  \| A y_{(1:k)} - h \|_2^2 \quad \forall y\in\R^d.
  \]
\end{theorem}
This theorem, combined with Theorem~\ref{thm:recon}, implies that OMP is
valid for matrices $A$ with $\mu(A) \leq 0.1/k$ and has output sparsity
$f(k) = 2k$.

{\bf $\ell_1$ algorithms.}
Basis Pursuit (BP) \cite{crt06} and its variants are based on finding the
minimum $\ell_1$-norm solution to a linear system.
While the basic form of BP is ill-suited for our application (it requires
the user to supply the amount of measurement error $\|Ay-h\|_2$), its more
advanced path-following or multi-stage variants may be valid~\cite{ehjt04}.

%

\section{Analysis}

\subsection{General Robustness Guarantees}

We now state our main regret transform bound, which follows immediately
from the definition of a valid reconstruction algorithm and linearity of
expectation.
\begin{theorem}[Regret Transform] \label{thm:regret}
Let $\rec$ be a valid reconstruction algorithm for $\{ \A_k : k \in \N\}$
and $\sperr:\N \times \R^d \to \R$.
Then there exists some constants $C_1$ and $C_2$ such that the following
holds.
Pick any $k \in \N$, $A \in \A_k$ with $m$ rows, and $H:\X\to\R^m$.
Let $F:\X\to\R^d$ be the composition of $\rec(k,A,\cdot)$ and $H$,
\emph{i.e.}~$F(x) = \rec(k,A,H(x))$.
Then
\begin{eqnarray*}
\E_x \| F(x) - \E[y|x] \|_2^2
& \leq &
C_1 \cdot \E_x \| H(x) - \E[Ay|x] \|_2^2
\ + \ C_2 \cdot \sperr(k, \E[y|x]).
\end{eqnarray*}
\end{theorem}
The simplicity of this theorem is a consequence of the careful composition
of the learned predictors with the reconstruction algorithm meeting the
formal specifications described above.

In order compare this regret bound with the bounds afforded by Sensitive
Error Correcting Output Codes (SECOC) \cite{lb05}, we need to relate
$\E_x\|H(x) - \E[Ay|x]\|_2^2$ to the average scaled mean-squared-error over
all induced regression problems; the error is scaled by the maximum
difference $L_i = \max_{y \in \Y} (Ay)_i - \min_y (Ay)_i$ between induced
labels:
\[ \bar{r} \ = \
\frac1m \sum_{i=1}^m \E_x \left( \frac{H(x)_i - \E[(Ay)_i|x]}{L_i}
\right)^2. \]
In $k$-sparse multi-label problems, we have $\Y = \{ y \in \{0,1\}^d :
\|y\|_0 \leq k \}$.
In these terms, SECOC can be tuned to yield
$\E_x\|F(x) - \E[y|x]\|_2^2 \ \leq \ 4k^2 \cdot \bar{r}$
for general $k$.

For now, ignore the sparsity error.
For simplicity, let $A \in \R^{m \times d}$ with entries chosen i.i.d.~from
the Bernoulli $B(1/2)$ distribution over $\{\pm1/\sqrt{m}\}$, where $m =
O(k \log d)$.
Then for any $k$-sparse $y$, we have $\|Ay\|_\infty \leq k/\sqrt{m}$, and
thus $L_i \leq 2k/\sqrt{m}$ for each $i$.
This gives the bound
\begin{equation*}
C_1 \cdot \E_x \| H(x) - \E[Ay|x] \|_2^2
\ \leq \ 4C_1 \cdot k^2 \cdot \bar{r},
\end{equation*}
which is within a constant factor of the guarantee afforded by SECOC.
Note that our reduction induces exponentially (in $d$) fewer subproblems
than SECOC.

Now we consider the sparsity error.
In the extreme case $m = d$, $\E[y|x]$ is allowed to be fully dense ($k =
d$) and $\sperr(k,\E[y|x]) = 0$.
When $m = O(k \log d) < d$, we potentially incur an extra penalty in
$\sperr(k,\E[y|x])$, which relates how far $\E[y|x]$ is from being
$k$-sparse.
For example, suppose $\E[y|x]$ has small $\ell_p$ norm for $0 \leq p < 2$.
Then even if $\E[y|x]$ has full support, the penalty will decrease
polynomially in $k \approx m / \log d$.

\subsection{Linear Prediction}

A danger of using generic reductions is that one might create a problem
instance that is even harder to solve than the original problem.
This is an oft cited issue with using output codes for multi-class problems.
In the case of linear prediction, however, the danger is mitigated, as we
now show.
Suppose, for instance, there is a perfect linear predictor of $\E[y|x]$,
\emph{i.e.}~$\E[y|x] = B^\top x$ for some $B \in \R^{p \times d}$ (here $\X
= \R^p$).
Then it is easy to see that $H = BA^\top$ is a perfect linear predictor of
$\E[Ay|x]$:
$$H^\top x \ = \ AB^\top x \ = \ A\E[y|x] \ = \ \E[Ay|x].$$
The following theorem generalizes this observation to imperfect linear
predictors for certain well-behaved $A$.

\begin{theorem} \label{thm:linear}
Suppose $\X \subset \R^p$.
Let $B \in \R^{p \times d}$ be a linear function with
\begin{equation*}
\E_x \left\| B^\top x - \mathbb{E}[y|x] \right\|_2^2 \ = \ \epsilon.
\end{equation*}
Let $A \in \mathbb{R}^{m \times d}$ have entries drawn i.i.d.~from
$N(0,1/m)$, and let $H = BA^\top$.
Then with high probability (over the choice of $A$),
\begin{equation*}
\E_x \|H^\top x - A\E[y|x]\|_2^2
\ \leq \
\left( 1 + O(1/\sqrt{m}) \right) \epsilon.
\end{equation*}
\end{theorem}
\begin{remark}
Similar guarantees can be proven for the Bernoulli-based matrices.
Note that $d$ does not appear in the bound, which is in contrast to the
expected spectral norm of $A$: roughly $1 + O(\sqrt{d/m})$.
\end{remark}
Theorem~\ref{thm:linear} implies that the errors of \emph{any} linear
predictor are not magnified much by the compression function.
So a good linear predictor for the original problem implies an
almost-as-good linear predictor for the induced problem.
Using this theorem together with known results about linear
prediction~\cite{kst08}, it is straightforward to derive sample
complexity bounds for achieving a given error relative to that of the best
linear predictor in some class.
The bound will depend polynomially in $k$ but only logarithmically in $d$.
This is cosmetically similar to learning bounds for feature-efficient
algorithms (\emph{e.g.}~\cite{ng04,kst08}) which are concerned with
sparsity in the weight vector, rather than in the output.

%

\section{Proofs} \label{sec:proofs}

\subsection{Proof of Theorem~\ref{thm:recon}}
Let $\ell = k + f(k)$, $y \in \R^d$, and assume without loss of generality
that $|y_1| \geq \ldots \geq |y_d|$.
We need to show that
$$\|\wh y - y\|_2^2
\ \leq \ C_1 \cdot \|Ay - h\|_2^2
\ + \ C_2 \cdot (\|\Delta\|_2^2 +
k^{-1} \|\Delta\|_1^2) $$
where $\Delta = y - y_{(1:k)}$.
Using the triangle inequality, the $(\ell,\delta)$-RIP of $A \in \A_k$,
and the hypothesis that $\|A\wh y - h\|_2^2 \leq C \|Ay_{(1:k)} - h\|_2^2$,
we have
\begin{eqnarray}
\|\wh y - y\|_2
& \leq & \|\wh y - y_{(1:k)}\|_2 + \|\Delta\|_2
\nonumber \\
& \leq & (1-\delta)^{-1/2} \|A\wh y - Ay_{(1:k)}\|_2 + \|\Delta\|_2
\nonumber \\
& \leq & (1-\delta)^{-1/2} \left(
\|A\wh y - h\|_2
+ \|h - Ay_{(1:k)}\|_2 \right) + \|\Delta\|_2
\nonumber \\
& \leq & (1-\delta)^{-1/2} (1 + \sqrt{C}) \|Ay_{(1:k)} - h\|_2 +
\|\Delta\|_2
\nonumber \\
& \leq & (1-\delta)^{-1/2} (1 + \sqrt{C}) \left( \|Ay-h\|_2 + \|A\Delta\|_2
\right) + \|\Delta\|_2.
\label{eq:here}
\end{eqnarray}
We need to relate $\|A\Delta\|_2$ to $\|\Delta\|_2$ and $\|\Delta\|_1$.
Write $\Delta = \sum_{i \geq 0} y_{J_i}$, where $J_i = \{ k + i\ell + 1,
\ldots, k + (i+1) \ell \}$ and $y_J \in \R^d$ is the vector whose $j$th
component is $y_j$ if $j \in J$ and is $0$ otherwise.
Note that each $y_{J_i}$ is $\ell$-sparse,
$\|y_{J_{i+1}}\|_1 \leq \|y_{J_i}\|_1$,
and $\|y_{J_{i+1}}\|_\infty \leq \ell^{-1} \|y_{J_i}\|_1$.
By H\"older's inequality,
$$
\|y_{J_{i+1}}\|_2
\ \leq \ (\|y_{J_{i+1}}\|_\infty \|y_{J_{i+1}}\|_1)^{1/2}
\ \leq \ (\ell^{-1} \|y_{J_i}\|_1^2)^{1/2}
\ = \ \ell^{-1/2} \|y_{J_i}\|_1,
$$
and so
$$
\sum_{i \geq 0} \|y_{J_i}\|_2
\ \leq \ \|y_{J_0}\|_2 + \sum_{i \geq 0} \|y_{J_{i+1}}\|_2
\ \leq \ \|y_{J_0}\|_2 + \ell^{-1/2} \sum_{i \geq 0} \|y_{J_i}\|_1
\ \leq \ \|\Delta\|_2 + \ell^{-1/2} \|\Delta\|_1.
$$
By the triangle inequality and the $(\ell,\delta)$-RIP of $A$,
we have
$$
\|A\Delta\|_2
\ \leq \ \sum_{i \geq 0} \|Ay_{J_i}\|_2
\ \leq \ \sum_{i \geq 0} (1+\delta)^{1/2} \|y_{J_i}\|_2
\ \leq \ (1+\delta)^{1/2} (\|\Delta\|_2 + \ell^{-1/2} \|\Delta\|_1).
$$
Combining this final inequality with \eqref{eq:here} gives
$$
\|\wh y - y\|_2
\ \leq \ C_0 \cdot \|Ay-h\|_2
\ + \ (1 + C_0(1+\delta)^{1/2})
\cdot (\|\Delta\|_2 + \ell^{-1/2} \|\Delta\|_1)
$$
where $C_0 = (1-\delta)^{-1/2}(1+\sqrt{C})$.
Now squaring both sides and simplifying using the fact $(x+y)^2 \leq 2x^2 +
2y^2$ concludes the proof.

\subsection{Proof of Theorem~\ref{thm:omp}}

We first begin with two simple lemmas.
\begin{lemma} \label{lem:small-progress}
Suppose OMP is run for $k$ iterations starting with $y^{(0)} = \vec 0$,
and produces intermediate solutions $y^{(1)}, y^{(2)}, \ldots, y^{(k)}$.
Then there exists some $0 \leq i < k$ such that if $j_i$ is the column
selected in step $i$, then $(a_{j_i}^\top (h - Ay^{(i)}))^2 \leq \|h\|_2^2
/ k$.
\end{lemma}
\begin{proof}
Let $r^{(i)} = h - Ay^{(i)}$.
Suppose column $j_i$ is added to $J$ in step $i$.
Let $\wt y^{(i+1)} = y^{(i)} + \alpha_i e_{j_i}$, where $\alpha_i =
a_{j_i}^\top r^{(i)}$ and $e_{j_i}$ is the $j_i$th elementary vector.
Then
\begin{align*}
\|r^{(i)}\|_2^2 - \|r^{(i+1)}\|_2^2
& \geq \|r^{(i)}\|_2^2 - \|h - A\wt y^{(i+1)}\|_2^2
\ = \ \|r^{(i)}\|_2^2 - \|h - A(y^{(i)} + \alpha_i e_{j_i})\|_2^2 \\
& = \|r^{(i)}\|_2^2 - \|r^{(i)} - \alpha_i a_{j_i}\|_2^2
\ = \ 2\alpha_i a_{j_i}^\top r^{(i)} - \alpha_i^2 \|a_{j_i}\|_2^2
\ = \ (a_{j_i}^\top r^{(i)})^2.
\end{align*}
Moreover, $\sum_{i=0}^{k-1} \|r^{(i)}\|_2^2 - \|r^{(i+1)}\|_2^2 =
\|r^{(0)}\|_2^2 - \|r^{(k)}\|_2^2 \leq \|h\|_2^2$, so there is some $i \in
\{0, 1, \ldots, k-1\}$ such that $(a_{j_i}^\top r^{(i)})^2 \leq
\|r^{(i)}\|_2^2 - \|r^{(i+1)}\|_2^2 \leq \|h\|_2^2 / k$.
\end{proof}

\begin{lemma} \label{lem:coherence}
If $y \in \R^d$ is $k$-sparse and $\mu(A) \leq \delta/(k-1)$, then
$\|Ay\|_2^2 \geq (1 - \delta) \|y\|_2^2$.
\end{lemma}
This result also appears in Appendix A1 of~\cite{det06}.
We reproduce the proof here.
\begin{proof}
Expanding $\|Ay\|_2^2$, we have
$$
\|Ay\|_2^2
\ = \ \sum_{i=1}^k \|a_i\|^2 y_i^2 + \sum_{i\neq j} y_i y_j (a_i^\top a_j)
\ \geq \ \|y\|_2^2 - \left|\sum_{i \neq j} y_iy_j (a_i^\top a_j)\right|,
$$
so we need to show this latter summation is at most $\delta \|y\|_2^2$.
Indeed,
\begin{align*}
\left|\sum_{i \neq j} y_iy_j (a_i^\top a_j) \right|
& \leq \sum_{i \neq j} |y_iy_j| |a_i^\top a_j|
&& \text{(triangle inequality)} \\
& \leq \mu(A) \sum_{i \neq j} |y_iy_j|
&& \text{(definition of coherence)} \\
& = \mu(A) \left( \sum_{i=1}^k \sum_{j=1}^k |y_i| |y_j| - \sum_{i=1}^k
y_i^2 \right) \\
& = \mu(A) \left( \left( \sum_{i=1}^k |y_i| \right)^2 - \|y\|_2^2 \right) \\
& \leq \mu(A) ( k \|y\|_2^2  - \|y\|_2^2 )
&& \text{(Cauchy-Schwarz)} \\
& = \mu(A) (k-1) \|y\|_2^2 \\
& \leq \delta \|y\|_2^2
&& \text{(assumption on $\mu(A)$)}
\end{align*}
which concludes the proof.
\end{proof}

We are now ready to prove Theorem~\ref{thm:omp}.
Without loss of generality, we assume that the columns of $A = [a_1 |
\ldots | a_d]$ are normalized (so $\|a_j\|_2=1$) and that the support of
$y$ is (some subset of) $\{1,\ldots,k\}$ (so $y$ is $k$-sparse).

In addition to the vector $\wh y$ returned by OMP and the vector $y$ we
want to compare to, we consider two other solution vectors:
\begin{itemize}
\item $y'$: a $(2k-1)$-sparse solution obtained by running up to $k-1$
iterations of OMP starting from $y$.
Lemma~\ref{lem:small-progress} implies that there exists such a vector $y'$
with the following property:
if $j^*$ is the column OMP would select when the current solution is $y'$,
then
\begin{eqnarray} \label{eq:omp-progress}
(a_{j^*}^\top (h - Ay'))^2 & \leq & \|h-Ay\|_2^2 / k.
\end{eqnarray}
Since $y'$ is obtained by starting with $y$, it can only have smaller
squared-error than $y$.
Without loss of generality, let the support of $y'$ be (some subset of) $\{
1,\ldots,2k\}$.

\item $\wh{y}'$: the actual solution produced by OMP (starting from $\vec
0$) just before OMP chooses a column $j \not\in \supp(y')$.
Note that if OMP never chooses a column $j \not\in \supp(y')$ within $2k$
steps, then $\|A\wh y - h\|_2^2 \leq \|A\wh y' - h\|_2^2 \leq \|Ay -
h\|_2^2$ and the theorem is proven.
Therefore we assume that this event does occurs and so $\wh y'$ is defined.
Since $\wh y'$ precedes the final solution $\wh y$ returned by OMP, it can
only have larger squared-error than $\wh y$.
\end{itemize}

We will bound $\|h - A\wh y\|_2$ as follows:
\begin{align*}
\|h - A\wh y\|_2
& \leq \|h - A\wh y'\|_2
&& \text{(since $\wh y'$ precedes $\wh y$)} \\
& \leq \|h - Ay'\|_2 + \|A(\wh y' - y')\|_2
&& \text{(triangle inequality)} \\
& \leq \|h - Ay\|_2 + \|A(\wh y' - y')\|_2.
&& \text{(since $y$ precedes $y'$)}
\end{align*}
We thus need to bound $\|A(\wh y' - y')\|_2$ in terms of $\|h-Ay\|_2$.

Let $\wh r = h - A\wh{y}'$ and $r = h - Ay'$.
Then
\begin{align*}
\|A(\wh y' - y')\|_2^2
& = (A\wh y' - Ay')^\top A(\wh y' - y') \\
& = (h - Ay')^\top A(\wh y' - y') - (h - A\wh y')^\top A(\wh y' - y') \\
& \leq \|h - Ay'\|_2 \|A(\wh y' - y')\|_2 + |(h - A\wh y')^\top A(\wh y' -
y')|
&& \text{(Cauchy-Schwarz)} \\
& = \|r\|_2 \|A(\wh y' - y')\|_2 + |\wh r^\top A(\wh y' - y')|.
\end{align*}
Using the fact $x \leq b\sqrt{x} + c \Rightarrow x \leq (4/3)(b^2 + c)$
(which in turn follows from the quadratic formula and the fact $2xy
\leq x^2 + y^2$), the above inequality implies
\begin{eqnarray} \label{eq:omp-bound1}
\frac34 \|A(\wh y' - y')\|_2^2
& \leq &
\|r\|_2^2 + |\wh r^\top A(\wh y' - y')|.
\end{eqnarray}
We now work on bounding the second term on the righthand side.
Let $j > 2k$ be the column chosen by OMP when the current solution is $\wh
y'$.
Then we have
\begin{eqnarray} \label{eq:omp-select}
| a_j^\top \wh r | & \geq & | a_\ell^\top \wh r | \quad \forall \ell \leq 2k.
\end{eqnarray}
Also, since $\wh y' - y'$ has support $\{1,\ldots,2k\}$, we have that
\begin{eqnarray} \label{eq:omp-support}
A(\wh y' - y') & = & A_{\{1:2k\}} (\wh y' - y')
\end{eqnarray}
where $A_{\{1:2k\}}$ is the same as $A$ except with zeros in all but the
first $2k$ columns.
Then,
\begin{align*}
|\wh r^\top A(\wh y' - y')|
& = |\wh r^\top A_{\{1:2k\}} (\wh y' - y')|
&& \text{(Equation~\eqref{eq:omp-support})} \\
& \leq \|\wh r^\top A_{\{1:2k\}}\|_\infty \|\wh y' - y'\|_1
&& \text{(H\"older's inequality)} \\
& \leq |a_j^\top \wh r| \|\wh y' - y'\|_1
&& \text{(Inequality~\eqref{eq:omp-select})} \\
& \leq \left( |a_j^\top r| + |a_j^\top A(\wh y' - y')| \right)
\|\wh y' - y'\|_1
&& \text{(triangle inequality)} \\
& \leq \left( |a_j^\top r| + \|a_j^\top A_{\{1:2k\}}\|_\infty \|\wh y' -
y'\|_1 \right) \|\wh y' - y'\|_1
&& \text{(Equation~\eqref{eq:omp-support} and H\"older)} \\
& \leq |a_j^\top r| \|\wh y' - y'\|_1 + \mu(A) \|\wh y' - y'\|_1^2
&& \text{(definition of coherence)} \\
& \leq \frac{5k}2 (a_j^\top r)^2 + \frac1{10k} \|\wh y' - y'\|_1^2
+ \mu(A) \|\wh y' - y'\|_1^2
&& \text{(since $xy \leq (x^2+y^2)/2$)} \\
& \leq \frac{5k}2 (a_j^\top r)^2 + \frac1{5k} \|\wh y' - y'\|_1^2
&& \text{(since $\mu(A) \leq 0.1/k$)} \\
& \leq \frac{5k}2 (a_j^\top r)^2 + \frac1{5k} (2k \|\wh y' - y'\|_2^2)
&& \text{(Cauchy-Schwarz)} \\
& \leq \frac{5k}2 (a_j^\top r)^2 + \frac12 \|A(\wh y' - y')\|_2^2.
&& \text{(Lemma~\ref{lem:coherence})}
\end{align*}
Continuing from Inequality~\eqref{eq:omp-bound1}, we have
\begin{eqnarray*}
\|A(\wh y' - y')\|_2^2
& \leq & 4\|r\|_2^2 + 10k (a_j^\top r)^2.
\end{eqnarray*}
Since $(a_j^\top r)^2 \leq (a_{j^*}^\top r)^2$, where $j^* \leq 2k$ is the
column that OMP would select when the current solution is $y'$, and since
$(a_{j^*}^\top r)^2 \leq \|h-Ay\|_2^2/k$ (by
Inequality~\eqref{eq:omp-progress}),
we have that
\begin{eqnarray*}
\|A(\wh y' - y')\|_2^2
& \leq & 4\|r\|_2^2 + 10 \|h-Ay\|_2^2 \\
& \leq & 14 \|h-Ay\|_2^2.
\end{eqnarray*}
Therefore,
\begin{eqnarray*}
\|h - A\wh y'\|_2
& \leq & (1+\sqrt{14}) \|h-Ay\|_2.
\end{eqnarray*}
Squaring both sides gives the conclusion.

\subsection{Proof of Theorem~\ref{thm:linear}}
We use the following Chernoff bound for sums of $\chi^2$ random
variables, a proof of which can be found in the Appendix A of
\cite{dasgupta00}.
\begin{lemma} \label{lem:chernoff}
Fix any $\lambda_1 \geq \ldots \geq \lambda_D > 0$, and let $X_1, \ldots,
X_D$ be i.i.d.~$\chi^2$ random variables with one degree of freedom.
Then
$\Pr[ \sum_{i=1}^D \lambda_i X_i > (1 + \gamma) \sum_{i=1}^D \lambda_i]
\leq
\exp(-(D\gamma^2/24) \cdot (\lambda/\lambda_1))
$
for any $0 < \gamma < 1$,
where $\lambda = (\lambda_1 + \ldots + \lambda_D) / D$.
\end{lemma}
Write $A = (1/\sqrt{m}) [\theta_1 | \cdots | \theta_m]^\top$, where each
$\theta_i$ is an independent $d$-dimensional Gaussian random vector
$N(0,I_d)$.
Define $v_x = B^\top x - \E[y|x]$ so $\epsilon \ = \ \E_x \|v_x\|_2^2$, and
assume without loss of generality that $v_x$ has full $d$-dimensional
support.
Using this definition and linearity of expectation, we have
$$ \E_x \|Av_x\|_2^2
\ = \ \frac1m \E_x \sum_{i=1}^m (\theta_i^\top v_x)^2
\ = \ \frac1m \sum_{i=1}^m \theta_i^\top (\E_x v_x v_x^\top) \theta_i.$$
Our goal is to show that this quantity is $(1 + O(1/\sqrt{m})) \epsilon$
with high probability.
Since $N(0,I_d)$ is rotationally invariant and $\E_x v_xv_x^\top$ is
symmetric and positive definite, we may assume $\E_x v_xv_x^\top$ is
diagonal and has eigenvalues $\lambda_1 \geq \ldots \geq \lambda_d > 0$.
Then, the above expression simplifies to
$$
\frac1m\sum_{i=1}^m \theta_i^\top (\E_x v_xv_x^\top) \theta_i
\ = \ \frac1m \sum_{i=1}^m \sum_{j=1}^d \lambda_j \theta_{ij}^2. $$
Each $\theta_{ij}^2$ is a $\chi^2$ random variable with one degree of
freedom, so $\E \theta_{ij}^2 = 1$.
Thus, the expected value of the above quantity is
$\sum_{j=1}^d \trace(\E_x v_x v_x^\top)
= \E_x \trace(v_x v_x^\top)
= \E_x \|v_x\|_2^2$.
Now applying Lemma~\ref{lem:chernoff}, with $D = md$ variables and $\lambda
= (\lambda_1 + \ldots + \lambda_d) / d$,
we have $\Pr[(1/m) \sum_{i,j} \lambda_j \theta_{ij}^2 > (1+t)\epsilon] \leq
\exp(-(mdt^2/24)(\lambda/\lambda_1))
\leq \exp(-mt^2/24)$
(using the fact $\lambda_1 \leq d \lambda$).
This bound is $\delta$ when $t = \sqrt{(24/m)\ln(1/\delta)}$.

\section{Experimental Validation}

\label{sec:experiments}

We conducted an empirical assessment of our proposed reduction on two
labeled data sets with large label spaces.
These experiments demonstrate the feasibility of our method -- a sanity
check that the reduction does in fact preserve learnability -- and compare
different compression and reconstruction options.

\subsection{Data}
\label{subsec:data}

\noindent {\bf Image data.\footnote{\texttt{http://hunch.net/$\sim$learning/ESP-ImageSet.tar.gz}}}
The first data set was collected by the ESP Game \cite{vad04}, an online
game in which players ultimately provide word tags for a diverse set of
web images.

The set contains nearly $68000$ images, with about $22000$ unique labels.
We retained just the $1000$ most frequent labels: the least frequent of
these occurs $39$ times in the data, and the most frequent occurs about
$12000$ times.
Each image contains about four labels on average.
We used half of the data for training and half for testing.

We represented each image as a bag-of-features vector in a manner similar
to \cite{mshv07}.
Specifically, we identified $1024$ representative SURF features points
\cite{betg08} from $10 \times 10$ gray-scale patches chosen randomly from
the training images; this partitions the space of image patches
(represented with SURF features) into Voronoi cells.
We then built a histogram for each image, counting the number of patches
that fall in each cell.

\noindent {\bf Text data.\footnote{\texttt{http://mlkd.csd.auth.gr/multilabel.html}}}
The second data set was collected by Tsoumakas et al.~\cite{tkv08} from
\texttt{del.icio.us}, a social bookmarking service in which users assign
descriptive textual tags to web pages.

The set contains about $16000$ labeled web page and $983$ unique labels.
The least frequent label occurs $21$ times and the most frequent occurs
almost $6500$ times.
Each web page is assigned $19$ labels on average.
Again, we used half the data for training and half for testing.

Each web page is represented as a boolean bag-of-words vector, with the
vocabulary chosen using a combination of frequency thresholding and
$\chi^2$ feature ranking. See \cite{tkv08} for details.

Each binary label vector (in both data sets) indicates the labels of the
corresponding data point.

\subsection{Output Sparsity}
\label{subsec:output-sparsity}

We first performed a bit of exploratory data analysis to get a sense of how
sparse the target in our data is.
We computed the
least-squares linear regressor $\wh B \in \R^{p \times d}$ on the training
data (without any output coding) and predicted the label probabilities $\wh
p(x) = \wh B^\top x$ on the test data (clipping values to the range $[0,1]$).
Using $\wh p(x)$ as a surrogate for the actual target $\E[y|x]$, we examined
the relative $\ell_2^2$ error of $\wh p$ and its best $k$-sparse approximation
$\epsilon(k,\wh p(x)) = \sum_{i = k+1}^d \wh p_{(i)}(x)^2/\|\wh
p(x)\|_2^2$, where $\wh p_{(1)}(x) \geq \ldots \geq \wh p_{(d)}(x)$.

Examining $\E_x \epsilon(k,\wh p(x))$ as a function of $k$, we saw that
in both the image and text data, the fall-off with $k$ is eventually
super-polynomial, but we are interested in the behavior for small $k$ where
it appears polynomial $k^{-r}$ for some $r$.
Around $k = 10$, we estimated an exponent of $0.50$ for the image data and
$0.55$ for the text data.
This is somewhat below the standard of what is considered sparse
(\emph{e.g.}~vectors with small $\ell_1$-norm show $k^{-1}$ decay).
Thus, we expect the reconstruction algorithms will have to contend with the
sparsity error of the target.

\subsection{Procedure}
\label{subsec:procedure}

We used least-squares linear regression as our base learning algorithm,
with no regularization on the image data and with $\ell_2$-regularization
with the text data ($\lambda = 0.01$) for numerical stability.
We did not attempt any parameter tuning.

The compression functions we used were generated by selecting $m$ random
rows of the $1024 \times 1024$ Hadamard matrix, for $m \in \{ 100, 200,
300, 400 \}$.
We also experimented with Gaussian matrices; these yielded similar but
uniformly worse results.

We tested the greedy and iterative reconstruction algorithms described
earlier (OMP, FoBa, and CoSaMP) as well as a path-following version of
Lasso based on LARS \cite{ehjt04}.
Each algorithm was used to recover a $k$-sparse label vector $\wh{y}^k$
from the predicted compressed label $H(x)$, for $k = 1, \ldots, 10$.
We measured the $\ell_2^2$ distance $\|\wh{y}^k-y\|_2^2$ of the prediction
to the true test label $y$.
In addition, we measured the precision of the predicted support at various
values of $k$ using the $10$-sparse label prediction.
That is, we ordered the coefficients of each $10$-sparse label prediction
$\wh{y}^{10}$ by magnitude, and measured the precision of predicting the
first $k$ coordinates $|\supp(\wh{y}^{10}_{(1:k)}) \cap \supp(y)|/k$.
Actually, for $k \geq 6$, we used $\wh{y}^{2k}$ instead of $\wh{y}^{10}$.

We used correlation decoding (CD) as a baseline method, as it is a standard
decoding method for ECOC approaches.
CD predicts using the top $k$ coordinates in $A^\top H(x)$, ordered by
magnitude.
For mean-squared-error comparisons, we used the least-squares approximation
of $H(x)$ using these $k$ columns of $A$.
Note that CD is not a valid reconstruction algorithm when $m < d$.

\subsection{Results}
\label{subsec:results}

\begin{figure}
\vskip 0.2in
\begin{center}
\includegraphics[width=\textwidth]{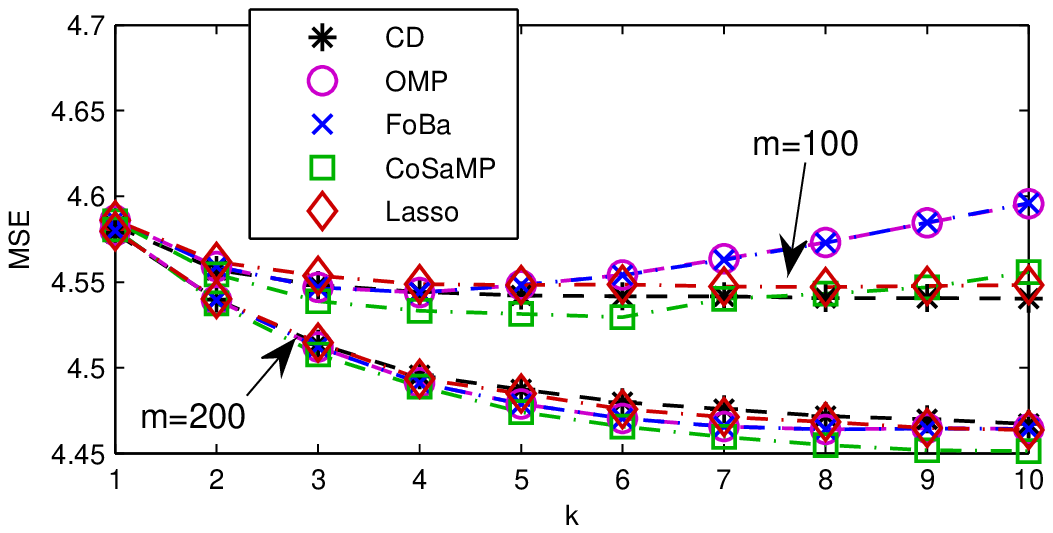} \\
\includegraphics[width=\textwidth]{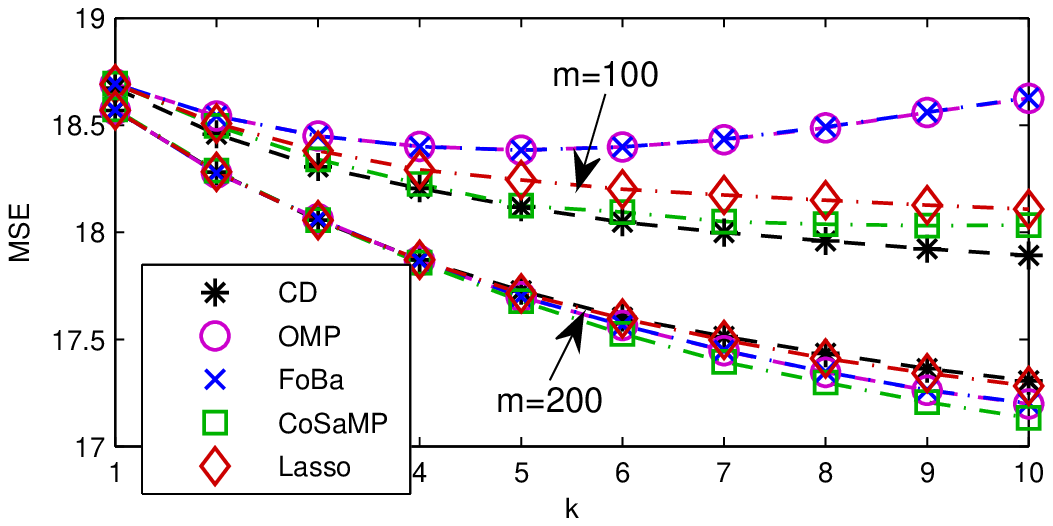}
\caption{Mean-squared-error versus output sparsity $k$, $m\in\{100,200\}$.
Top: image data. Bottom: text data.
In each plot:
the top set of lines corresponds to $m=100$,
and the bottom set to $m=200$.}
\label{fig:mse}
\end{center}
\vskip -0.2in
\end{figure}

\begin{figure}
\vskip 0.2in
\begin{center}
\includegraphics[width=\textwidth]{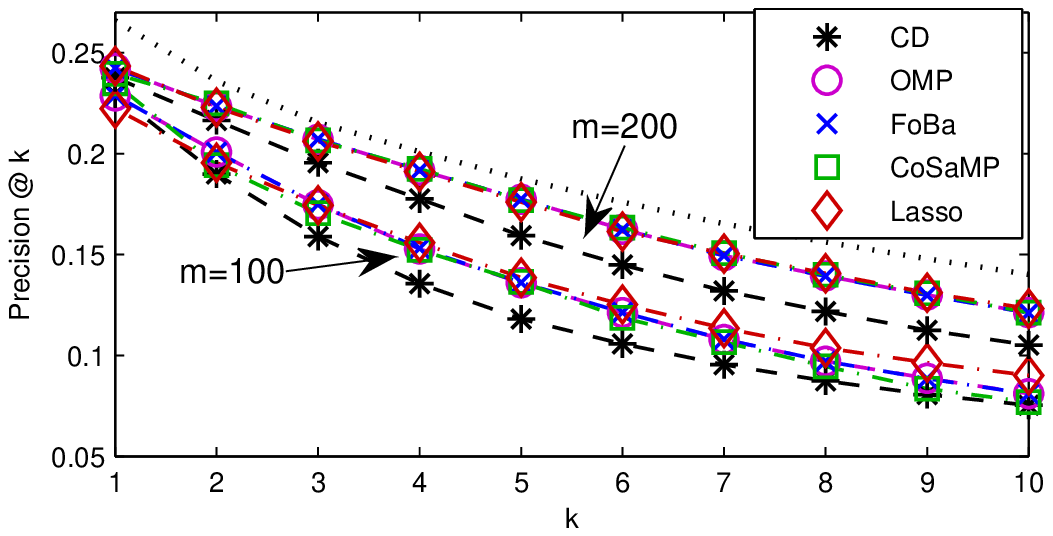} \\
\includegraphics[width=\textwidth]{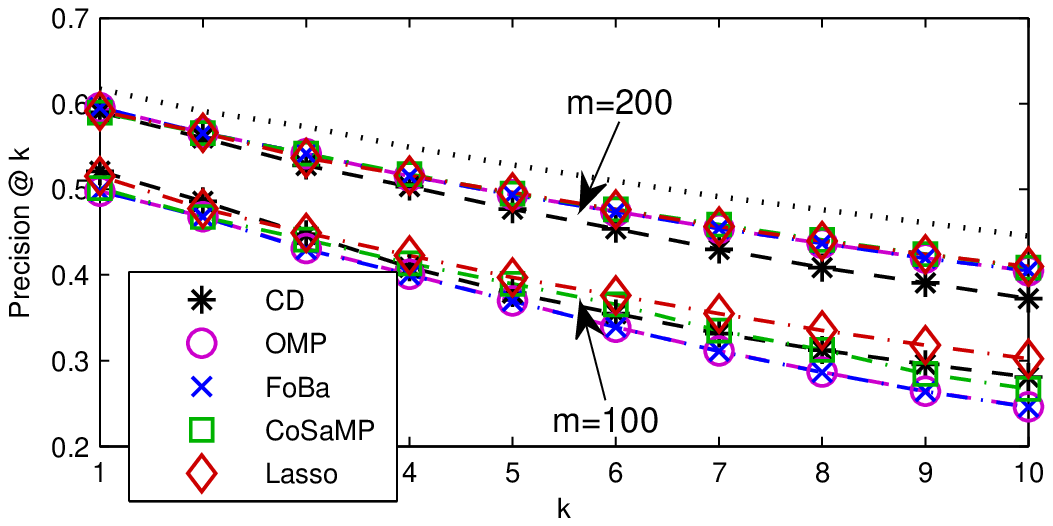}
\caption{Mean precision-at-$k$ versus output sparsity $k$, $m\in\{100,200\}$.
Top: image data. Bottom: text data.
In each plot:
the top black unadorned line is one-against-all ($m=1024$),
the middle set of lines corresponds to $m=200$,
and the bottom set to $m=100$.  }
\label{fig:pak}
\end{center}
\vskip -0.2in
\end{figure}

As expected, the performance of the reduction, using any reconstruction
algorithm, improves as the number of induced subproblems $m$ is increased
(see Figures~\ref{fig:mse} and \ref{fig:pak}; at $m=300,400$, the
precision-at-$k$ is nearly the same as one-against-all,
\emph{i.e.}~$m=1024$).
When $m$ is small and $A \not\in \A_K$, the reconstruction algorithm cannot
reliably choose $k \geq K$ coordinates, so its performance may degrade
after this point by over-fitting.
But when the compression function $A$ is in $\A_K$ for a sufficiently
large $K$, then the squared-error decreases as the output sparsity $k$
increases up to $K$.
Note the fact that precision-at-$k$ decreases as $k$ increases is expected,
as fewer data will have at least $k$ correct labels.

All of the reconstruction algorithms at least match or out-performed the
baseline on the mean-squared-error criterion, except when $m = 100$.
When $A$ has few rows, (1) $A \in \A_K$ only for very small $K$, and (2)
many of its columns will have significant correlation.
In this case, when choosing $k > K$ columns, it is better to choose
correlated columns to avoid over-fitting.
Both OMP and FoBa explicitly avoid this and thus do not fare well; but
CoSaMP, Lasso, and CD do allow selecting correlated columns and thus
perform better in this regime.

The results for precision-at-$k$ are similar to that of mean-squared-error,
except that choosing correlated columns does not necessarily help in the
small $m$ regime.
This is because the extra correlated columns need not correspond to
accurate label coordinates.

In summary, the experiments demonstrate the feasibility and robustness of
our reduction method for two natural multi-label prediction tasks.
They show that predictions of relatively few compressed labels are
sufficient to recover an accurate sparse label vector, and as our theory
suggests, the robustness of the reconstruction algorithms is a key factor
in their success.

\subsubsection*{References}
{\def\section*#1{}\small \bibliography{multi_compressed}
\bibliographystyle{unsrt}}

\end{document}